 \documentclass[pmlr,twocolumn,10pt]{jmlr} 

\usepackage{booktabs}
\usepackage{siunitx}

\usepackage{hyperref}

\theorembodyfont{\upshape}
\theoremheaderfont{\scshape}
\theorempostheader{:}
\theoremsep{\newline}

\jmlrvolume{LEAVE UNSET}
\jmlryear{2023}
\jmlrsubmitted{LEAVE UNSET}
\jmlrpublished{LEAVE UNSET}
\jmlrworkshop{Machine Learning for Health (ML4H) 2023} 

 \title[NCL-SM]{Introducing NCL-SM: A Fully Annotated Dataset of Images from Human Skeletal Muscle Biopsies }

\author{%
\Name{Atif Khan}\Email{a.khan21@newcastle.ac.uk}\\
\Name{Conor Lawless \textsuperscript{*}} \Email{conor.lawless@newcastle.ac.uk}\\
\Name{Amy Vincent} \Email{amy.vincent@newcastle.ac.uk}\\
\Name{Charlotte Warren} \Email{Charlotte.Warren@newcastle.ac.uk}\\
\Name{Valeria Di Leo} \Email{Valeria.Di-Leo@newcastle.ac.uk}\\
\Name{Tiago Gomes} \Email{Tiago.Gomes@newcastle.ac.uk}\\
\Name{A. Stephen McGough}  \Email{stephen.mcgough@newcastle.ac.uk}\\
\addr Wellcome Centre for Mitochondrial Research, Translational and Clinical Research Institute, Faculty of Medical Sciences, Newcastle University, UK; School of Computing, Newcastle University, UK
\newline * Corresponding Author 
}

\begin{document}

\maketitle

\begin{abstract}
Single cell analysis of skeletal muscle (SM) tissue is a fundamental tool for understanding many neuromuscular disorders. For this analysis to be reliable and reproducible, identification of individual fibres within microscopy images (segmentation) of SM tissue should be precise. There is currently no tool or pipeline that makes automatic and precise segmentation and curation of images of SM tissue cross-sections possible. Biomedical scientists in this field rely on custom tools and general machine learning (ML) models, both followed by labour intensive and subjective manual interventions to get the segmentation right. We believe that automated, precise, reproducible segmentation is possible by training ML models. However, there are currently no good quality, publicly available annotated imaging datasets available for ML model training. In this paper we release NCL-SM: a high quality bioimaging dataset of 46 human tissue sections from healthy control subjects and from patients with genetically diagnosed muscle pathology. These images include $>$ 50k manually segmented muscle fibres (myofibres). In addition we also curated high quality myofibres and annotated reasons for rejecting low quality myofibres and regions in SM tissue images, making this data completely ready for downstream analysis. This, we believe, will pave the way for development of a fully automatic pipeline that identifies individual myofibres within images of tissue sections and, in particular, also classifies individual myofibres that are fit for further analysis.   
\end{abstract}

\section{Introduction}
\label{sec:intro}
Microscopy and cytometry-based techniques allow us to observe protein expressions within individual cells, when a single cell segmentation approach is used. For many biological and disease processes, this is the most appropriate spatial scale for understanding mechanisms.  Further, the spatial arrangement of cells of different classes within tissues \textit{in vitro} is often informative about biology and disease pathology. 
There are many diseases affecting SM tissue, including Amyotrophic Lateral Sclerosis  \citep{Wales2011SeminarSclerosis}, Multiple Sclerosis \citep{Filippi2018MultipleSclerosis}, Muscular Dystrophy \citep{Bushby2010ReviewManagement} and a wide range of mitochondrial diseases \citep{Morrison2009NeuromuscularDiseases}. The dataset we present here is collected from healthy human control subjects and from patients suffering from genetically diagnosed muscle pathology, including mitochondrial diseases. Mitochondrial diseases are individually uncommon but are collectively the most common metabolic disorder affecting 1 in 5,000 people \citep{Ng2016MitochondrialManagement}. They can cause severe disabilities and adversely affect the life expectancy of patients \citep{Barends2016}. Mitochondrial disease pathology is complex and highly heterogeneous \citep{Ng2016MitochondrialManagement}. Some of the latest approaches to classifying mitochondrial diseases and quantifying disease severity are based on the analysis of single myofibre protein expression profiles and clinical information from patients \citep{Alston2017}.  These same approaches are currently being successfully applied to other types of muscle pathology \citep{DiLeo2023JournalX-xx}.  
\newline Any analysis of SM images at the single myofibre level requires 1) precise segmentation of individual myofibres, as regions close to the myofibre membrane are known to exhibit differential features \citep{Vincent2018SubcellularMuscle}. 2) removal of myofibres damaged by freezing that might occur in the process of storage and thawing, as the subcellular patterns in protein expression, or indeed per-cell mean expression, will be impacted by the technical artefact, detracting from the target biology. 3) removal of SM myofibres that are not sliced in transverse sections or are partially sectioned, as the presence of such myofibres does not allow for a standard or uniform comparison across all the myofibres in a tissue and 4) removal of folded tissue. Tissue can fold in on itself during tissue handling and slide preparation. Such folding artificially amplifies apparent protein expression in these regions and is again a technical artefact, not related to target biology.  

Currently most single-myofibre SM analysis is carried out using custom built semi-automatic pipelines like mitocyto \citep{Warren2020}, using general image analysis tools like Ilastik \citep{Berg2019Ilastik:Analysis} or cellprofiler \citep{Carpenter2006CellProfiler:Phenotypes}, or using vanilla ML models like stardist \citep{Schmidt2018} or cellpose \citep{Stringer2020Cellpose:Segmentation}. None of these approaches produce segmentation quality required for analysis of SM in our experience. Custom built pipelines like mitocyto are used more often than general ML models (which are not trained on SM data) as these require relatively fewer corrections than general ML models. To improve the segmentation quality and remove compromised myofibres and SM regions, biomedical scientists spend hours manually correcting the issues in a tissue section before doing downstream quantitative analysis. This can be an inefficient use of scientist's time but also the corrections are subjective and not reproducible.

The main barrier to the development of a suitable ML tool/model for fully automatic segmentation and curation of SM myofibres is the lack of any high quality, manually segmented and curated SM myofibre image data on which to train appropriate ML models. In this paper we make available a high quality fully manually segmented and mostly manually curated SM imaging dataset collected using two different imaging technologies: Imaging Mass Cytometry (IMC) and ImmunoFluorescence (IF) amounting to 50,434 myofibres in 46 tissue sections, 30,794 of which are classed as `analysable', 18,102 classed as `not-analysable-due-to-shape', 1,538 myofibres classed as `not-analysable-due-to-freezing-damage' and 405 annotations of folded tissue regions. We describe the Newcastle Skeletal Muscle (NCL-SM) dataset, introduce quantitative quality assessment metrics relevant to single-myofibre SM tissue segmentation and demonstrate the annotation quality of our ground truth manual annotations in terms of duplicate human-to-human annotations for quality assurance.

The main contributions of this paper are:
\begin{itemize}
    \item Publishing a high quality, manually segmented set of IMC and IF images of human skeletal muscle tissue, curated by expert biomedical scientists: the NCL-SM dataset. We provide NCL-SM\footnote{\url{https://doi.org/10.25405/data.ncl.24125391}} and our code\footnote{\url{www.github.com/atifkhanncl/NCL-SM} } that evaluates annotation quality and classify non-transverse sliced myofibres.
    \item Introducing quantitative quality assessment metrics for SM myofibre segmentation. 
 
\end{itemize}

\section{Tissue Collection and Image Curation}
\label{sec:review}

The Wellcome Centre for Mitochondrial Research (WCMR)\footnote{\url{www.newcastle-mitochondria.com/}} is one of the leading institutes conducting research into mitochondrial diseases worldwide, and we have an unparalleled repository of clinical data and tissue from healthy control subjects and patients with genetically diagnosed muscle pathology, largely mitochondrial disease  patients with mitochondrial myopathy. The NCL-SM dataset includes images of 46 tissue sections that capture spatial variation in protein expression within tissue (including within myofibres). We use microscopy-based techniques i.e. IF and advanced protein expression measurement techniques like IMC that allow us to observe the spatial variation in the expression of up to 40 proteins in tissue simultaneously.

The assessment process before ethical approval for collection of SM tissue from human donors is granted, is rigorous and muscle tissue donations can be painful.  As a result, SM tissue collections are scarce and valuable, requiring  available SM tissue sections to be used efficiently. To achieve this, biomedical scientists in this field follow processes to preserve and efficiently use the available samples throughout the research cycle. This also applies to the imaging data captured from these tissues i.e. within the images it is critical to efficiently curate all usable myofibres. This can only be achieved by precisely segmenting all available SM fibres and curating the subset that are fit for further analysis. Currently, this is a heavily manual process as there does not exist any automatic tools or pipeline that can perform this. 

Machine learning has made some great strides in image segmentation, object detection and image classification \citep{Minaee2022ImageSurvey, Greenwald2022Whole-cellLearning, Schmidt2018,Stringer2020Cellpose:Segmentation}. However, leveraging ML tools is only possible with good quality, large training datasets. 
To our knowledge there does not exist any such dataset for SM fibre segmentation or classification that is publicly available. Making this dataset available and clearly defining the challenge involved in myofibre segmentation and curating the analysis worthy myofibres is a crucial first step towards development of an automatic tool for this problem. 

High quality annotated datasets are critical for development of relevant ML models or pipelines. This is evident since the early days of modern ML with datasets such as MNIST \citep{Deng2012TheResearch}, COCO \citep{Lin2014LNCSContext} and more recently SA-1B \citep{Kirillov2023SegmentAnything} enabling the construction of some seminal ML models like  ResNet \citep{He2015DeepRecognition}, VGG \citep{Simonyan}, vision transformer \citep{Dosovitskiy2020ANSCALE} and SAM \citep{Kirillov2023SegmentAnything}.

\section{Capturing images}
The following is the sequential process of collecting this data.

\begin{table*}[]
    \caption{All annotation counts in NCL-SM dataset. In the table TS, AM, NTM,FAM and FR stands for Tissue Section, Analysable Myofibre, Non-Transverse Myofibre, Freezing Artefact Myofibre and Folded regions respectively. }
    \centering
    \begin{tabular}[width=1\textwidth]{||c c c c c c c||} 
     \hline
     Imaging Technique & TS Count & Myofibre Count & AM Count & NTM Count & FAM Count & FR Count\\  
     \hline\hline
     IMC & 27 & 22,979 & 14,841 & 7,358 & 780 & 84\\ 
     \hline
    IF & 19 & 27,455 & 15,953 & 10,744 & 758 & 321 \\ 
     \hline
     Total & 46 & 50,434 & 30,794 & 18,102 & 1,538 & 405 \\ 
     \hline
    \end{tabular}
    \label{tab:all dataset details}
\end{table*}

\subsection{Biopies}
Following ethical approval from the Newcastle and North Tyneside Local Research Ethics Committee and informed consent from control and patient subjects, biopsies of SM were collected from  patient and healthy control subjects. However, Ethical approval for this project was not required as it consisted of a re-analysis of existing image data. This original data was obtained from approved studies on samples from Newcastle Brain Tissue Resource (Approval Ref:2021031) and Newcastle Biobank (Application Ref: 042). 

\subsection{IMC}
\begin{figure*}[htbp]
\includegraphics[width=1\textwidth]{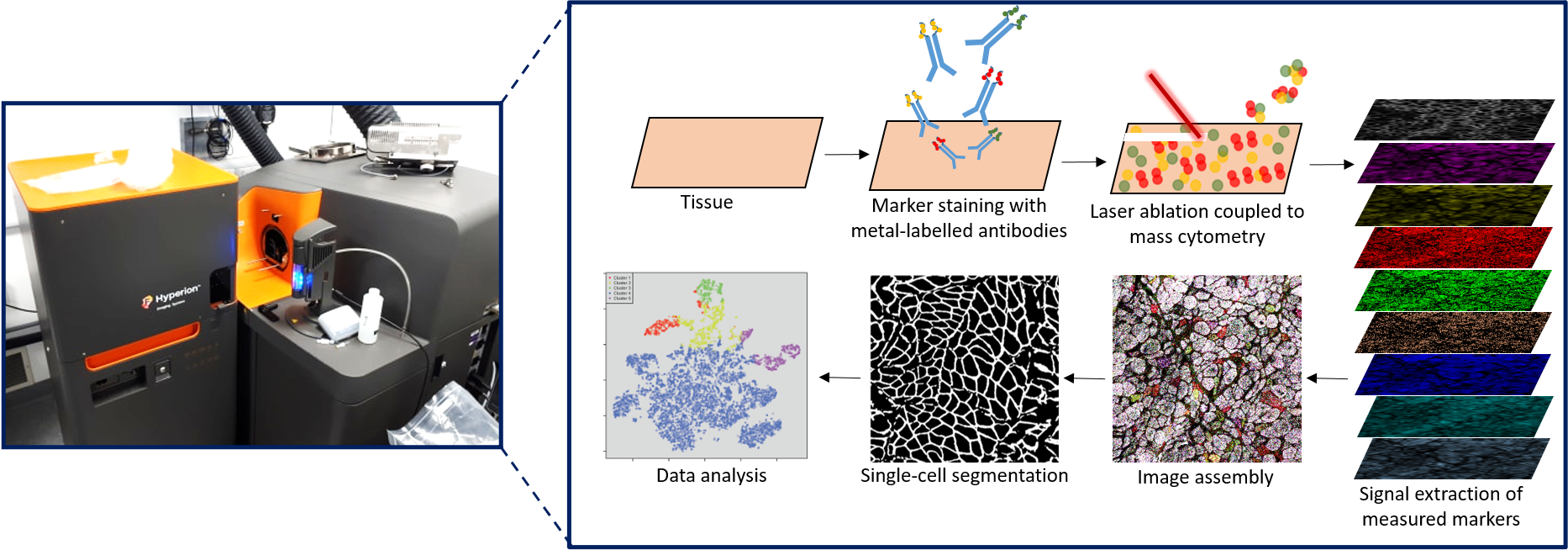}
\caption{\textbf{Imaging Mass Cytometry Experimental Procedure} During IMC myofibres are stained with a panel of antibodies conjugated to heavy metals. Tissue (made up of myofibres) is scanned by a pulsed laser which ablates a spot of tissue section. The tissue is vaporised on each laser shot and enters the mass cytometer where the relative concentration of heavy metals can be quantified. These measurements are later combined as pseudo images, one for each metal/protein, where location and intensity of each pixel correspond to the amount of metal isotopes at each spot. Myofibres are segmented in these images for conducting single myofibre data analysis. Figure adapted from \citet{Giesen2014HighlyCytometry}}. 
\label{Fig:IMC process figure}
 \end{figure*}

Imaging mass cytometry is an imaging technique to simultaneously observe the expression of multiple proteins in a tissue section as described in Figure \ref{Fig:IMC process figure}.

\paragraph{NCL-SM images}
The 46 tissue section images in NCL-SM are made by arranging greyscale images of a cell membrane protein marker i.e. dystrophin and of a mitochondrial mass protein marker VDAC1 into an RGB image where R = membrane protein marker, G =  mass protein marker and B = 0. Each channel of the images are contrast stretched (5 to 95 percentile) to enhance visual quality for the segmentation task but raw images without contrast stretching are also included in NCL-SM.

\subsection{IF}
 Immunofluorescence imaging is another technology that uses the antibodies to their antigen to target fluorescent dyes to specific protein targets within a cell/myofibres. It allows the capture of high resolution and high bit-depth microscopy images, but only up to 5 protein targets can be observed simultaneously \citep{Im2019ChapterStaining}.

\subsection{Myofibre Segmentation}
\label{myofibre_segment}
The protocol for fibre segmentation was i) include all areas within a myofibre that had mitochondrial mass signal, ii) exclude any areas within a myofibre that had myofibre membrane signal and iii) prioritise signal from within myofibre when membrane signal is weak.  Point iii) is necessary because noise is common in these types of data resulting in some overlapping mass and membrane pixels.  In such scenarios, we consider mitochondrial mass signal within a myofibre to be the most reliable indicator of myofibre morphology. 
We employed annotation specialists from Gamaed\footnote{\url{www.gamaed.com/}} working under the close oversight of expert biomedical scientists from WCMR to segment each myofibre in all 46 tissue sections, amounting to 50,434 myofibres. Using the online Apeer platform\footnote{\url{www.apeer.com/}} for all manual annotations. All segmentation went through rigorous visual inspection and a number of random myofibres in the data were segmented separately by expert biomedical scientists for quality assurance (QA). This is further discussed in Section \ref{Sec: QA}.

\begin{figure}[htbp]
\centerline{\includegraphics[width=.5\textwidth]{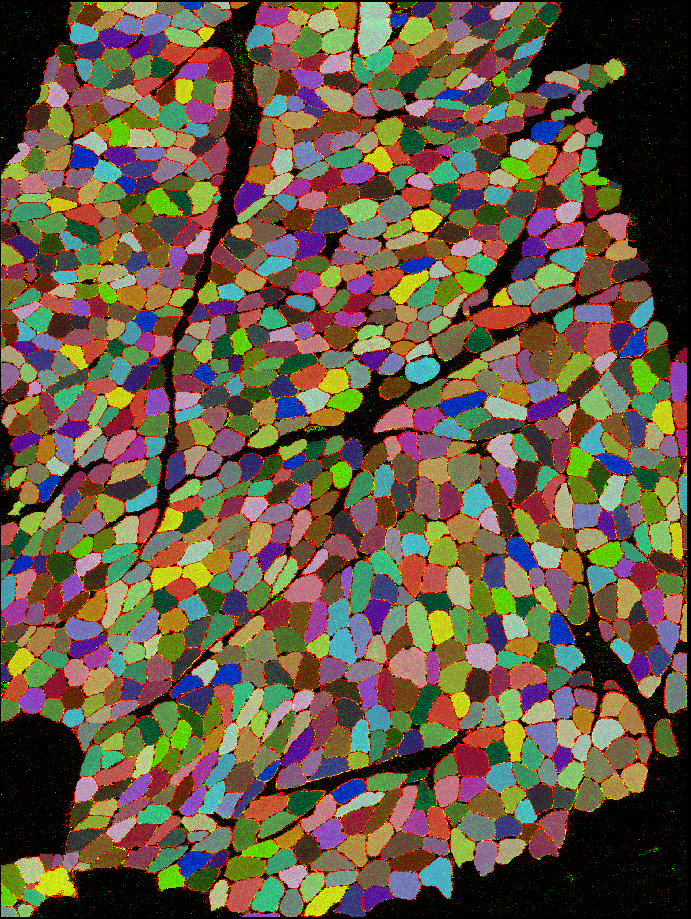}}
\caption{\textbf{Typical Manual Segmentation of a SM Tissue Section}:  An IMC SM tissue section image from subject `P17'. Consists of 1,068 myofibres manually segmented following the protocol i.e. include all myofibre mass and exclude membrane }
\label{Fig:Typical myofibre segmentation}
\end{figure}

\subsection{Freezing Artefact Myofibre Classification}
\begin{figure}[htbp]
\centerline{\includegraphics[width=.5\textwidth]{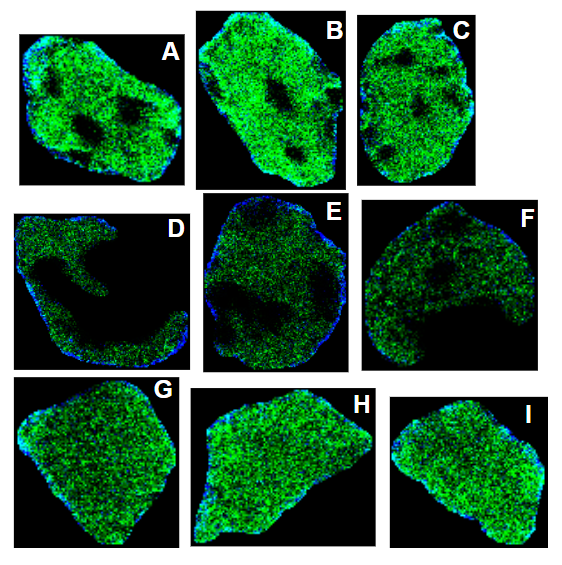}}
\caption{\textbf{Identifying Myofibre Freezing Artefacts} All myofibres from section P02. On top row myofibres A (id:190), B (id:138) and C (id:257) are typical freezing damaged myofibres resulting in leopard spots pattern, on middle row are the partial myofibres D (id:448), E (id:117) and F (id:398) that are damaged by freezing, and on bottom row are myofibres G (id:303), H (id:415) and I (id:288) that are without freezing defects.}
\label{Fig:Typical freezing damaged fibres}
\end{figure}

The protocol for identifying myofibres with freezing artefacts is i) look for leopard spot pattern within myofibres that are typical of freezing damage and ii) look for partial myofibres i.e. large part of myofibres missing as a result of freezing. As demonstrated in Figure \ref{Fig:Typical freezing damaged fibres}.

The freezing artefact classification annotation was duplicated by two experts from WCMR and any disagreement was resolved by discussion between a panel of experts. This resulted in 1,538 SM fibres with freezing artefacts where both annotators were in agreement.  All annotations which differed between annotators were reviewed and resolved by a second expert after discussion. 

\subsection{Non-Transverse Sliced Myofibre Classification}
\begin{figure}[htbp]
\centerline{\includegraphics[width=.5\textwidth]{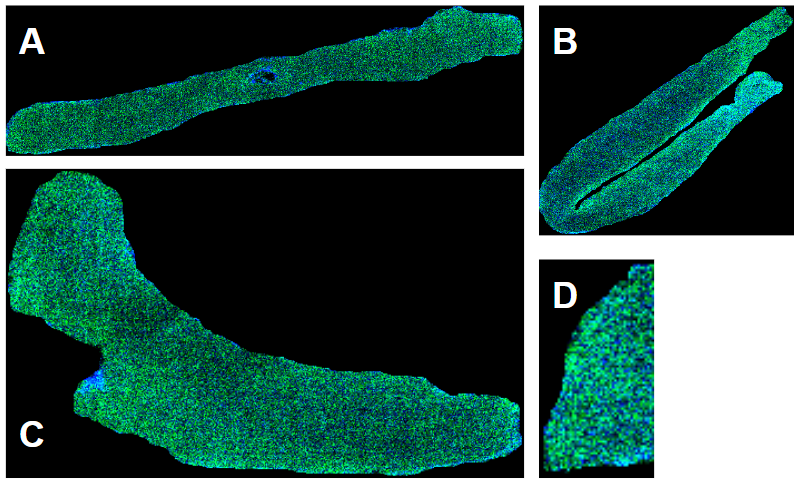}}
\caption{\textbf{Typical Non-Transverse Sliced Myofibres}: Myofibres from tissue section `C04'. `A' (ID: 39 ) is a typical elongated myofibre, `B' is myofibre (ID:6) with unusual convexity, `C' is myofibre (ID:24) of large area and `D' is a partial myofibre (ID:12) on the border of the image }
\label{Fig:Typical Non-Transverse sliced  fibres}
\end{figure}

The protocol for identifying non-transverse sliced myofibres was to look for i)  myofibres with skewed aspect ratio e.g. elongated ii) all myofibres at the border of image: these are partial observations iii) segmented objects which are too small or too big and iv) myofibres with unusual convexity. As demonstrated in Figure \ref{Fig:Typical Non-Transverse sliced  fibres},
for annotating non-transverse sliced fibres we employed the following approach 1) two experts working together identified up to 1,500 such myofibres in the data 2) using these 1,500 myofibres, thresholds for area, convexity and aspect ratio were calculated. 3) these thresholds and a function to detect any myofibre on the edge were then applied on whole data, resulting in two classes of myofibres. 4) finally both classes of myofibres were rigorously visually inspected to detect and correct any mis-classification. These resulted in 18,102  non-transverse sliced myofibres (NTM).  

\subsection{Folded Tissue Segmentation}
Folded tissue regions were segmented by an expert biomedical scientist resulting in annotations for 405 different tissue regions affecting 37 out of 46 sections.

\begin{figure}[htbp]
\centerline{\includegraphics[width=.5\textwidth]{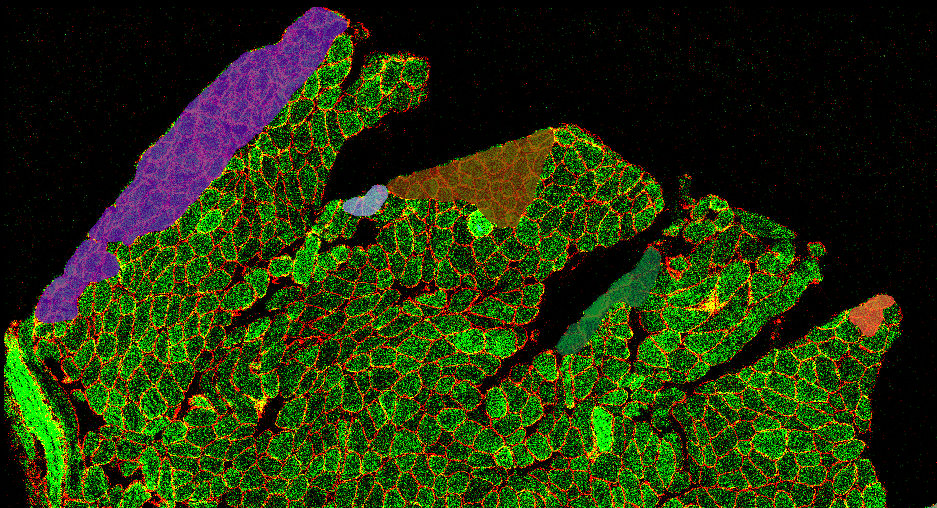}}
\caption{\textbf{Typical SM Tissue Folding}  Segmentation of folded regions in a tissue section P06}
\label{Fig:Typical SM Tissue Folding}
\end{figure}

\section{Quality Evaluation Metrics and Measurements}
\label{Sec: QA}
\begin{table*}[]
    \caption{ Reported are the annotation quality metrics' values for QA human-to-human annotation comparison as mention in Section \ref{myofibre_segment}. In the table MF-A, $r \textsubscript{AoB}$, $r \textsubscript{AiB}$, $\overline{IoU}$, ( A\%(IoU $>$0.80), A\%(IoU $>$0.90),  A\%(IoU $>$0.95)), QA-IMC, QA-IF stands for `Myofibres Assessed', `Myofibre Mass Missed Correlation', `Myofibre Membrane Included Correlation', 'Mean IoU', ('Accuracy in terms of \% of myofibres meeting IoU thershold of 0.8, 0.9 and 0.95'), `QA for IMC images' and `QA for IF images' respectively. }
    \label{tab:benchmark_ann}
    \centering
    \begin{tabular}[width=1\textwidth]{||c c c c c c c c ||} 
    
     \hline
     \newline
     Annotations & MF-A  & $r \textsubscript{AoB}$&  $r \textsubscript{AiB}$ & $\overline{IoU}$ & A\%(IoU $>$0.80) & A\%(IoU $>$0.90) & A\%(IoU $>$0.95) \\  
     \hline\hline
     QA-IMC & 53 & 0.99&  0.77 & 0.96 & 100 & 100& 77.4  \\ 
     \hline
     QA-IF & 23 & 0.92 &  0.94 & 0.96 & 100 & 100& 74  \\ 
     \hline
    
    \end{tabular}
\end{table*}
\subsection{Assessment Metrics}

\begin{figure}[htbp]
\centerline{\includegraphics[width=.5\textwidth]{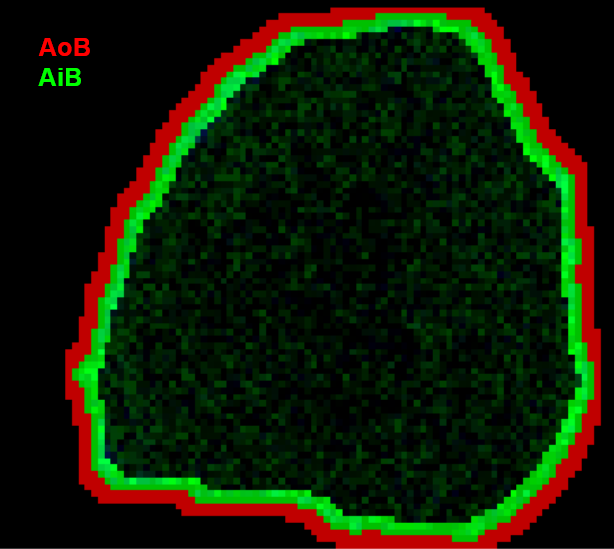}}
\caption{\textbf{Area Near the Membrane}  IMC image of a fibre (ID:723) from tissue section `P17' illustrating Area outside the Border (AoB) and Area inside the  Border(AiB) on either side of myofibre annotated border identified by eroding and dilating the border using a 5x5 and 9x9 pixel kernels for IMC and IF images respectively.}
\label{Fig:Areas_near_myofibre_border}
\end{figure}

Intersection over Union (IoU) is widely used evaluation metric to measure the quality of annotation/segmentation in computer vision tasks. But for myofibre segmentation examining IoU of each myofibre alone will not reveal important aspects about segmentation quality i.e. the area missed or included matter as emphasised in Sections  \ref{sec:intro} and \ref{myofibre_segment}, in other words we want any automatic pipeline to have high accuracy segmenting areas on either side of the border of each myofibre as illustrated in Figure \ref{Fig:Areas_near_myofibre_border}. To measure this, we developed two quantitative metrics  `Myofibre Mass Missed Correlation' and `Myofibre Membrane Included Correlation' along with the aformentioned IoU. All three metrics are defined below:

\begin{itemize}

\item \textbf{Myofibre Mass Missed Correlation}: For annotator x, y this is the Pearson correlation of myofibre mass pixels missed in Area outside the myofibre Border (AoB) between two duplicate annotations across all myofibres.

\item \textbf{Myofibre Membrane Included Correlation}: For annotator x, y this is Pearson correlation of myofibre membrane pixels included in Area inside the myofibre border (AiB) between two duplicate annotations across all myofibres assessed.
\item \textbf{IoU}: This is defined as intersection of overlapping pixels divided by union of all pixels between two annotations of the same myofibre.

\end{itemize}

\subsection{Assessment Measurements}

As mentioned in Section \ref{myofibre_segment}, for QA during the manual segmentation, experts at WCMR routinely annotated random myofibres in various sections to make sure the quality of annotations by specialists from Gamaed remain high quality. The observations are listed in Table \ref{tab:benchmark_ann}. As these are comparison of duplicate manual annotations, we believe this should be used as benchmark for any automatic segmentation tool or pipeline.

\section{Conclusion and Future Work}
In this paper we release NCL-SM dataset, a high quality dataset of $>$ 50k annotated SM myofibre segmentation with curated myofibre classified between fit for analysis SM fibres and not. We define assessment of segmentation quality for single myofibre SM analysis, introducing the metrics relevant for this. We demonstrate the high quality of annotation of NCL-SM in terms of relevant metrics. We define the challenges involved in segmenting SM myofibres for possible automatic ML solution. We believe this dataset will enable development of novel ML solutions to address the problem of automatic segmentation of SM fibres and classification of individual myofibres that are fit for downstream analysis. 

For future work we shall expand NCL-SM to include more high quality annotated data not only from our centre but establish a process in place by which others can add their data which meet the quality standards of NCL-SM.

\acks{This work was supported by the EPSRC Centre for Doctoral Training in Cloud Computing for Big Data and Wellcome Centre for Mitochondrial Research. AEV is in receipt of a Newcastle University Academic Track Fellowship. Annotation work was carried out by an independent, specialist data labelling company: \href{https://www.gamaed.com/}{Gamaed}.}

\bibliography{references}

\end{document}